\documentclass{article}

\usepackage{microtype}
\usepackage{graphicx}
\usepackage{subfigure}
\usepackage{booktabs} % for professional tables

\usepackage{hyperref}

\usepackage[accepted]{icml2024}

\usepackage{amsmath}
\usepackage{amssymb}
\usepackage{mathtools}
\usepackage{amsthm}

\usepackage{bbm}
\usepackage{wrapfig}
\usepackage{graphicx}
\usepackage{multirow}

\usepackage[capitalize,noabbrev]{cleveref}

\theoremstyle{plain}

\theoremstyle{definition}

\theoremstyle{remark}

\usepackage[textsize=tiny]{todonotes}

\icmltitlerunning{DirecT2V: Large Language Models are Frame-Level Directors for Zero-Shot Text-to-Video Generation}

\begin{document}

\twocolumn[
\icmltitle{DirecT2V: Large Language Models are Frame-Level Directors for Zero-Shot Text-to-Video Generation}

% It is OKAY to include author information, even for blind
% submissions: the style file will automatically remove it for you
% unless you've provided the [accepted] option to the icml2024
% package.

% List of affiliations: The first argument should be a (short)
% identifier you will use later to specify author affiliations
% Academic affiliations should list Department, University, City, Region, Country
% Industry affiliations should list Company, City, Region, Country

% You can specify symbols, otherwise they are numbered in order.
% Ideally, you should not use this facility. Affiliations will be numbered
% in order of appearance and this is the preferred way.
\icmlsetsymbol{equal}{*}

\begin{icmlauthorlist}
\icmlauthor{Susung Hong}{univ}
\icmlauthor{Junyoung Seo}{univ}
\icmlauthor{Heeseong Shin}{univ}
\icmlauthor{Sunghwan Hong}{univ}
\icmlauthor{Seungryong Kim}{univ}
\end{icmlauthorlist}

\icmlaffiliation{univ}{Korea Univeristy, Seoul, Korea}

\icmlcorrespondingauthor{Seungryong Kim}{seungryong\_kim@korea.ac.kr}

% You may provide any keywords that you
% find helpful for describing your paper; these are used to populate
% the "keywords" metadata in the PDF but will not be shown in the document
\icmlkeywords{Machine Learning}

\vskip 0.3in
]

% this must go after the closing bracket ] following \twocolumn[ ...

% This command actually creates the footnote in the first column
% listing the affiliations and the copyright notice.
% The command takes one argument, which is text to display at the start of the footnote.
% The \icmlEqualContribution command is standard text for equal contribution.
% Remove it (just {}) if you do not need this facility.

\printAffiliationsAndNotice{}  % leave blank if no need to mention equal contribution
% \printAffiliationsAndNotice{\icmlEqualContribution} % otherwise use the standard text.

\begin{abstract}

In the paradigm of AI-generated content (AIGC), there has been increasing attention to transferring knowledge from pre-trained text-to-image (T2I) models to text-to-video (T2V) generation. Despite their effectiveness, these frameworks face challenges in maintaining consistent narratives and handling shifts in scene composition or object placement from a single abstract user prompt. Exploring the ability of large language models (LLMs) to generate time-dependent, frame-by-frame prompts, this paper introduces a new framework, dubbed DirecT2V. DirecT2V leverages instruction-tuned LLMs as directors, enabling the inclusion of time-varying content and facilitating consistent video generation. To maintain temporal consistency and prevent mapping the value to a different object, we equip a diffusion model with a novel value mapping method and dual-softmax filtering, which do not require any additional training. The experimental results validate the effectiveness of our framework in producing visually coherent and storyful videos from abstract user prompts, successfully addressing the challenges of zero-shot video generation.

\end{abstract}

\begin{figure*}[t]
\centering
\includegraphics[width=1.0\linewidth]{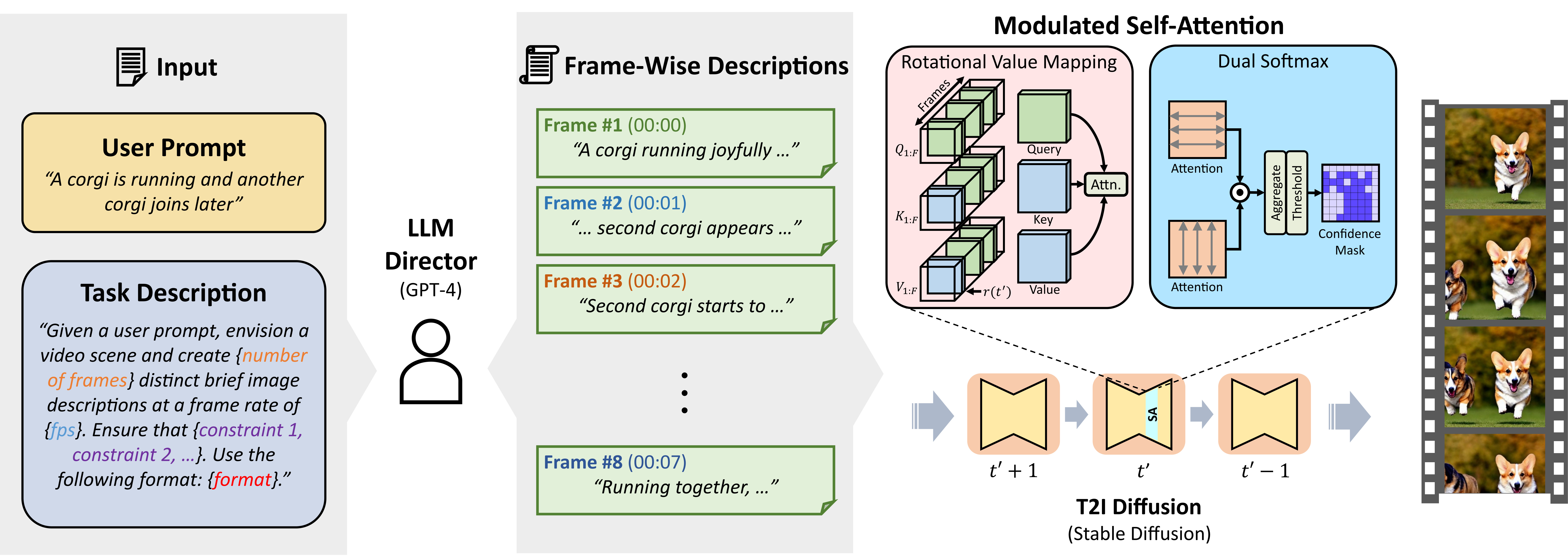}
\vspace{-10pt}
\caption{\textbf{Overall pipeline of our DirecT2V framework.} DirecT2V consists of two parts: directing an abstract user prompt with an LLM director (\textit{e.g.}, GPT-4~\cite{openai2023gpt4}) and video generation with a T2I diffusion (\textit{e.g.}, Stable Diffusion~\cite{rombach2022high}) equipped with modulated self-attention.}
\vspace{-15pt}
\label{fig:method}
\end{figure*}

\section{Introduction}

Within the paradigm of AI-generated content (AIGC), there has been increasing attention in expanding the capabilities of pre-trained text-to-image (T2I) models to text-to-video (T2V) generation~\cite{khachatryan2023text2video,singer2022make,hong2022cogvideo,singer2022make,blattmann2023align,zhou2022magicvideo}. One notable advancement in this area is the Text2Video-Zero (T2V-Z) framework, which introduced a fine-tuning-free approach utilizing a pre-trained T2I diffusion model~\cite{rombach2022high,saharia2022photorealistic} for generating videos from text descriptions~\cite{khachatryan2023text2video} in a zero-shot manner. Additionally, several other studies~\cite{khachatryan2023text2video,wu2022tune,qi2023fatezero}  have focused on enhancing temporal consistency in existing text-to-image diffusion models by redesigning the self-attention module, enabling video generation without the need for further training. These methods have successfully reduced the requirement for expensive fine-tuning processes, saving both time and resources while ensuring accessibility.

Although these methods have proven effective~\cite{khachatryan2023text2video,wu2022tune,qi2023fatezero}, they are not without drawbacks. One significant challenge is the use of a single user prompt to condition all frames, which may struggle to maintain consistent narratives and varying contexts over time. In contrast to images, which can be described by one or a few sentences, videos contain sequences of time-varying actions and contexts, requiring much more descriptive information. This challenge is not fully addressed in such works, as they all assume a single prompt conditions the entire video. Consequently, the limited ability to comprehend the temporal dynamics of complex actions from a single prompt, which provides only abstract information, can result in videos that overlook important aspects such as motions, actions, or events.

To address these limitations, we devise a novel framework that leverages instruction-tuned large language models (LLMs)~\cite{ouyang2022training,wei2021finetuned}, such as GPT-4~\cite{openai2023gpt4} and PaLM2~\cite{google2023palm2}, for generating frame-by-frame descriptions in video creation from a single abstract user prompt. Starting from the analysis of LLMs' ability to generate time-varying frame-level directions, we propose a method called DirecT2V, which enables zero-shot video creation by utilizing carefully designed task prompts tailored for instruction-tuned LLMs. To this end, we use LLM directors to divide user inputs into separate prompts for each frame, essentially separating static and dynamic elements within the user prompts. This separation allows for the integration of these components into text-to-image models, enabling the inclusion of time-varying content, which was previously unattainable, and facilitating consistent video generation.

Although the complemented prompt may help, creating temporally cohesive and visually captivating videos from a text-to-image model remains extremely challenging due to the stochasticity of diffusion models~\cite{ho2020denoising,song2020score,karras2022elucidating}. Therefore, in addition to high-level narrative synthesized through frame-level descriptions, we allow frame interactions to achieve temporal coherence and flexibility between frames. Specifically, we propose methods that adaptively and seamlessly integrate into the self-attention mechanism in T2I diffusion models, namely value mapping and dual-softmax filtering. Based on diffusion timesteps, value mapping selects an arbitrary frame among all frames and propagates the self-attention value to others. To make the mapping more reliable, dual-softmax filtering obtains and leverages the confidence masks from the self-attention layers, thereby eliminating unreliable mappings between frames.

To evaluate our framework, we present extensive experimental results that demonstrate the effectiveness of the proposed methods in addressing the challenges of zero-shot video generation from abstract user prompts. Empirical results validate the effectiveness of the DirecT2V framework in producing visually coherent and consistent videos from abstract user prompts.

\section{Related Work}

\begin{figure*}[t]
\centering
\includegraphics[width=0.8\linewidth]{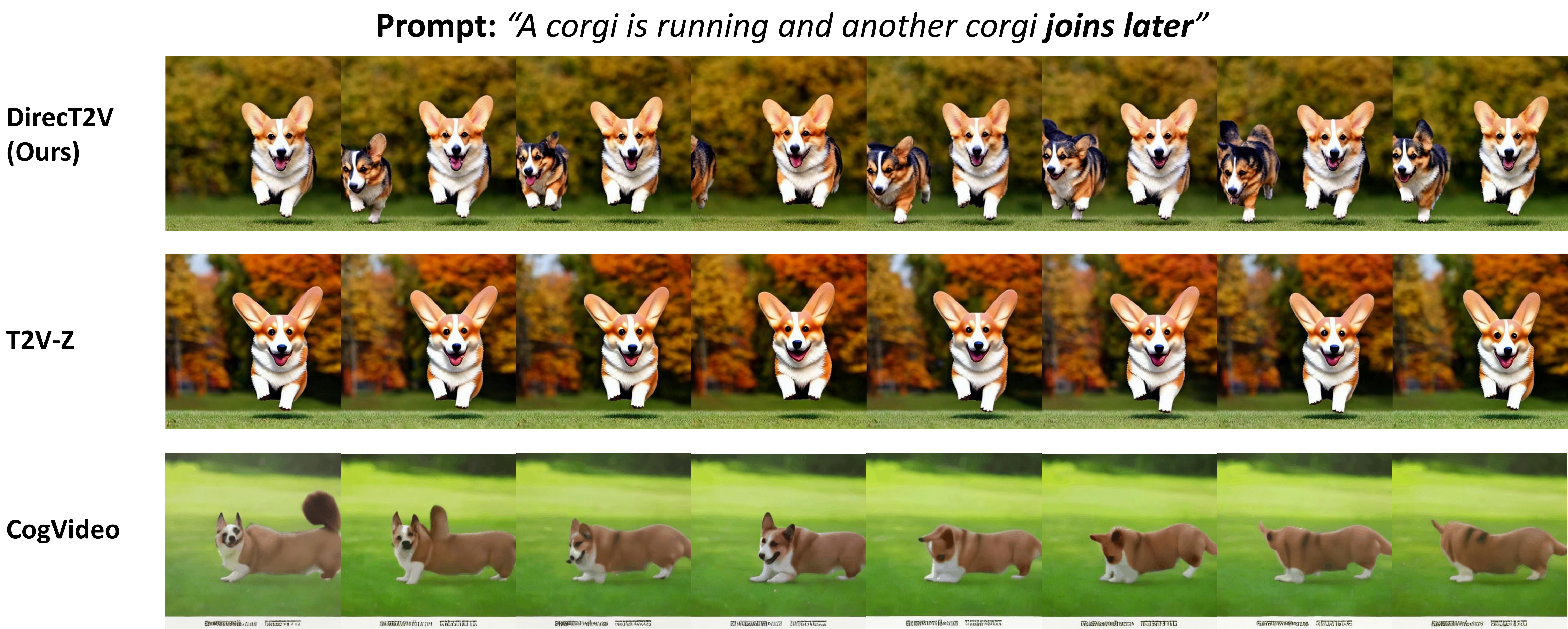}
\vspace{-10pt}
\caption{\textbf{Zero-shot video generation.} DirecT2V, using LLMs as frame-level directors, enables zero-shot narrative text-to-video generation, while current zero-shot (\textit{e.g.}, Text2Video-Zero~\cite{khachatryan2023text2video}) or tuned (\textit{e.g.}, CogVideo~\cite{hong2022cogvideo}) baselines do not contain high-level stories, \textit{e.g.}, joining of the second corgi.}
\vspace{-15pt}
\label{fig:teaser}
\end{figure*}

\paragraph{Incorporating large language models.}

Large language models (LLMs), such as GPT-3~\cite{brown2020language}, PaLM~\cite{chowdhery2022palm}, and BLOOM~\cite{scao2022bloom}, have been shown to be effective in a wide range of tasks, \textit{e.g.}, decision making~\cite{li2022pre}, program synthesis~\cite{austin2021program}, and prompt engineering~\cite{zhou2022large}. Notably, their zero-shot capabilities have demonstrated strong generalization power that almost resembles the linguistic ability of humans. By transferring such knowledge, numerous methods~\cite{saharia2022photorealistic,li2022blip,koizumi2020audio,brown2020language} have excelled at even tasks involving different modalities, \textit{i.e.}, audio, text, and images. Specifically, a recently introduced technique called instruction-finetuning, achievable via supervision or RLHF~\cite{stiennon2020learning,christiano2017deep}, enabled accurate manipulation of LLMs that aligns with human intent.
Another line of works, including~\cite{brooks2023instructpix2pix,hao2022optimizing}, have proposed to combine pre-trained language models with diffusion-based generative models, aiming to generate prompts that produce more reliable results. However, the capability of LLMs to recognize the time variation of a video scene or to generate time-varying prompts for a single user prompt has rarely been explored.
\vspace{-5pt}

\paragraph{Text-to-video generation.}

In the stream of research on AI-generated content (AIGC), text-to-video generation has been receiving considerable attention as a forefront research area, exploring various methodologies to generate videos from textual inputs. Among them, some methods employ autoregressive transformers or diffusion processes~\cite{wu2022nuwa,villegas2022phenaki,ho2022imagen,ho2022cascaded}. N\"UWA~\cite{wu2022nuwa} introduces a 3D transformer~\cite{vaswani2017attention}-based encoder-decoder framework and aims to tackle various tasks, including text-to-video generation, while Phenaki~\cite{villegas2022phenaki} presents a bidirectional masked transformer for a video creation from arbitrary-length text prompt sequences. Similarly, Imagen Video~\cite{ho2022imagen} leverages diffusion models for cascading pipeline~\cite{ho2022cascaded} and introduces a freamework to spatial and temporal super-resolution.

Notably, a recent trend is that owing to remarkable generation ability of large-scale text-to-image models, numerous methods attempted to transfer their knowledge and even extend to other tasks, including text-to-video generation. CogVideo~\cite{hong2022cogvideo} builds upon CogView2~\cite{ding2022cogview2}, a text-to-image model, and employs a multi-frame-rate hierarchical training strategy, encouraging text and video alignment. Make-a-video~\cite{singer2022make} tackles T2V task more efficiently by using the synthetic data for self-supervision. Another line of works~\cite{ho2022video,zhou2022magicvideo} that exploit LDM~\cite{rombach2022high} enables high-resolution video generation by introducing temporal tuning technique that efficiently fine-tunes the parameters. Taking a step forward, Text2Video-Zero (T2V-Z)~\cite{khachatryan2023text2video} introduces tuning-free zero-shot video generation without requiring intensive training or large-scale video datasets.  

Although aforementioned works may synthesize temporally consistent and high fidelity videos, it is notable that a single user prompt is responsible for the actions in all the frames in a video, making the output videos lacking story. As illustrated in Fig.~\ref{fig:teaser}, the time-dependent dynamics are often disregarded and only limitedly expressed. In this work, we obtain time-varying prompts from a single user prompt using an instruction-tuned LLM~\cite{ouyang2022training,openai2023gpt4,google2023palm2,wei2021finetuned}, and use them to synthesize videos that successfully capture both static and dynamic components. Regarding discussions on concurrent works, we direct the readers to the appendix.

\begin{figure*}[t]
\centering
\includegraphics[width=0.8\linewidth]{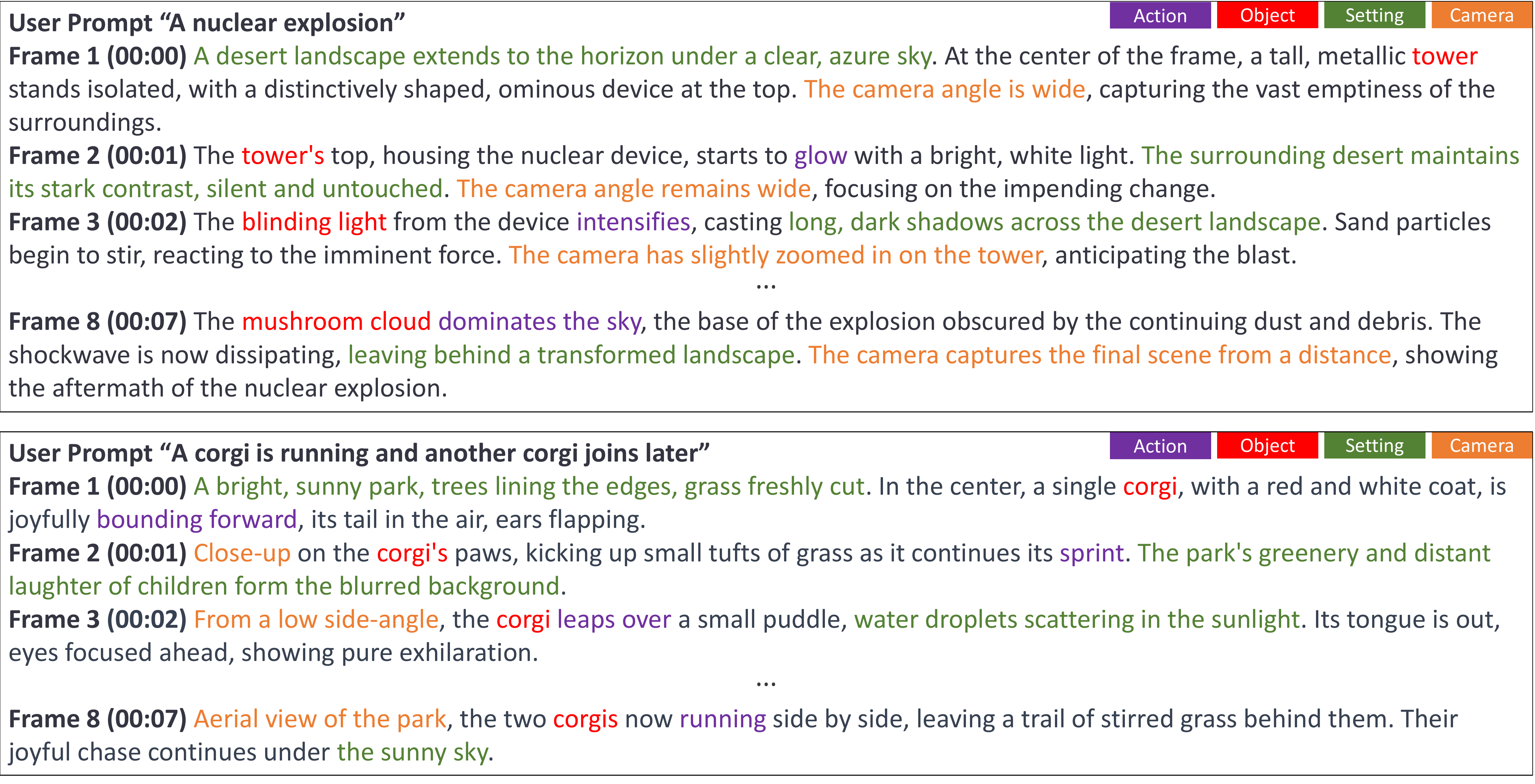}
\vspace{-10pt}
\caption{\textbf{Examples of frame-level directing with LLM.} Given an abstract user prompt, our LLM director outputs frame-wise prompts that complement the initial prompts with insufficient information. For complete instructions, see the appendix.}
\vspace{-15pt}
\label{fig:llm}
\end{figure*}

\section{Method}

\subsection{Preliminaries}

Numerous works in text-to-image (T2I) field, which include GLIDE~\cite{nichol2021glide}, Dall-E 2~\cite{ramesh2022hierarchical}, latent diffusion models (LDM)~\cite{rombach2022high} and Imagen~\cite{saharia2022photorealistic}, have been actively employing diffusion models for their high-fidelity generation. In this section, we first explain details of LDM~\cite{rombach2022high} whose methods are adopted in Stable Diffusion.

LDM~\cite{rombach2022high} is a diffusion model that performs the forward and reverse process within the latent space of an autoencoder denoted as $D(E(\cdot))$, where $E(\cdot)$ and $D(\cdot)$ symbolize the encoder and decoder, respectively. Given an input image $x\in \mathbb{R}^{H \times W \times 3}$ and its latent tensor $z_0 := E(x) \in \mathbb{R}^{h \times w \times c}$ where  $h < H$ and $w < W$, during the forward process, Gaussian noise is progressively added to $z_0$ as follows:
\begin{equation}
q(z_t|z_{t-1}) = \mathcal{N}(z_t; \sqrt{1 - \beta_t} z_{t-1}, \beta_t I),
\end{equation}
where $t = 1, \dots, T$, and $q(z_t|z_{t-1})$ denotes the conditional density of $z_t$ given $z_{t-1}$, and ${\beta_t}$ for all $t$'s are hyperparameters that defines the noise schedule. The forward process is repeated until the initial signal $z_0$ is entirely obscured, yielding $z_T \sim \mathcal{N}(0, I)$. The objective of the diffusion models is then to learn the reverse process defined as:
\begin{equation}
p_\theta(z_{t-1}|z_t) = \mathcal{N}(z_{t-1}; \mu_\theta(z_t, t), \Sigma_\theta(z_t, t)),
\end{equation}
where $t = T, \dots, 1$. This enables the discovering of a valid signal $z_0$ from the standard Gaussian noise $z_T$. To sample from $p_\theta(z_{t-1}|z_t)$, LDM~\cite{rombach2022high} instead predicts the reparametrization $\epsilon_\theta(z_t, t)$. To achieve text-conditioned image sampling, the text embedding of a user prompt $\omega$ is conditioned along the intermediate features via the cross-attention layers~\cite{vaswani2017attention,rombach2022high}, resulting in the conditional prediction term $\epsilon_\theta(z_t, t, \omega)$, and classifier-free guidance~\cite{ho2021classifier} is leveraged for better alignment to user prompts.

\subsection{Motivation}
\label{sec:motivation}

Text-to-video generation is a challenging task, particularly in a zero-shot setting where video-specific priors like temporal consistency and motion realism cannot be learned due to the absence of video data during training. Moreover, the generated video should align with the provided text descriptions, capturing the narrative essence of the scene. Recently, T2V-Z~\cite{khachatryan2023text2video} has tackled this task by lifting text-to-image (T2I) models to text-to-video (T2V) generation.

Despite its noticeable performance, T2V-Z is confined to having simple motions based on a specific linear movement and stochastic perturbation from the diffusion process. In specific, motions are introduced by translating the latent tensor in x-/y-axis in the early process of generation. Not only this is distant from motion realism, it also hinders the alignment between the video and the text description. For example, given the occasional presence or absence of objects within frames (\textit{e.g.}, \textit{``joining''} or \textit{``exiting''}) and the representation of changing object states (\textit{e.g.}, \textit{``jump''} or \textit{``explode''}). Therefore, when generating videos that align with user prompts, there is often a need to handle dynamic contents within the prompt, subject to narrative (plot) and temporal consistency.  

To overcome these challenges in generating full video frames from a single prompt for the text-to-video generation task, we find that pre-trained large language models (LLMs) can effectively transform abstract user prompts into frame-by-frame image descriptions. For extensive analyses of this ability, we refer readers to the appendix. This strategy addresses the multifaceted challenges of temporal consistency and visual quality. The components that can be captured by LLMs include: \textbf{actions}, \textbf{object descriptions}, \textbf{contextual information}, \textbf{camera angles and movements}, and \textbf{plot or storyline}. We show that each element is well expressed and it is illustrated in Fig.~\ref{fig:llm}. In Sec.~\ref{sec:prompting}, we detail how the proposed method satisfies the aforementioned items. 

Although this approach generates high-level frame-by-frame descriptions, plain LDM~\cite{rombach2022high}, originally intended for random image generation, generally cannot generate a video. An aspect that LDM lacks, primarily due to its stochastic nature, but which we aim to include, is temporal consistency. Without such consideration, crucial in video generation tasks, total discontinuity between adjacent frames occurs. As a remedy, recent studies have shown that sharing the same self-attention matrices within the U-Net of diffusion models can achieve object rigidity~\cite{khachatryan2023text2video,wu2022tune,qi2023fatezero}. Nonetheless, this approach heavily relies on prior attention maps to compute attention at arbitrary time steps, enforcing unchanging contexts, as exemplified in Fig.~\ref{fig:teaser}. To address this, we explore methods to interact frames within the video, enhancing flexibility and overall quality of generated videos while preserving temporal consistency.

\begin{figure*}[t]
\centering
\includegraphics[width=1.0\linewidth]{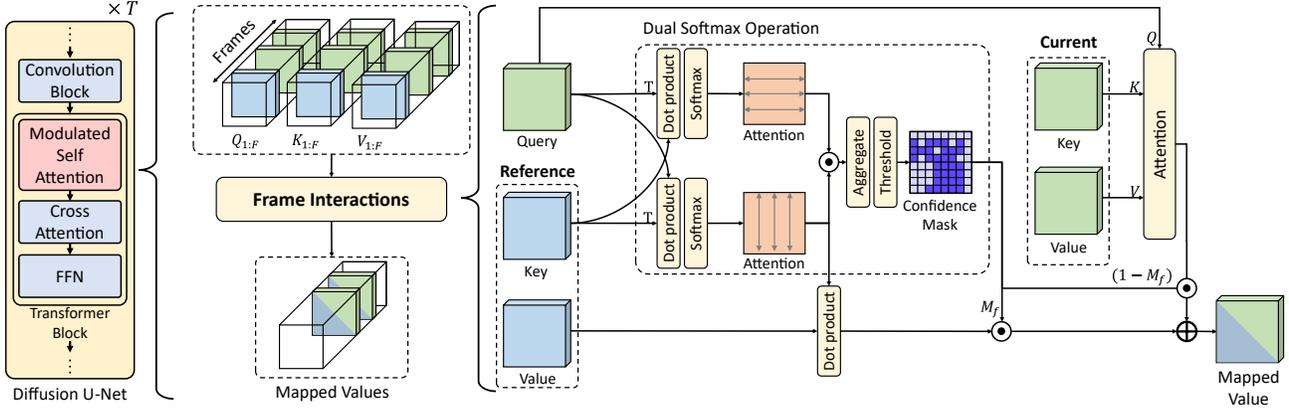}
\vspace{-10pt}
\caption{\textbf{The proposed frame interaction mechanism.} Within our augmented self-attention layers in a diffusion U-Net~\cite{ronneberger2015u}, value mapping and dual-softmax are introduced to ensure temporal consistency and reduce unreliable matching between frames, respectively.}
\vspace{-10pt}
\label{fig:attention-ours}
\end{figure*}

\subsection{Frame-wise directing with LLM}
\label{sec:prompting}

In order to effectively leverage instruction-tuned LLMs~\cite{ouyang2022training,openai2023gpt4,google2023palm2,wei2021finetuned} for video generation, we claim that it is crucial to take into account the narrative consistency, in other words, the \textit{storyline} reflected to the video. To achieve this goal, we propose a dynamic prompting strategy, in order to grant controllability over the desired attributes of the video without hurting the narratives. For this, we provide LLMs with a user prompt indicating the narrative for the overall scene with a task description to ensure the continuity of the narrative and controllability of the various attributes of the video, such as the number of frames and frames per second (FPS).

As shown in Fig.~\ref{fig:method}, given a prompt \textit{``A corgi is running, and another corgi joins later,''} we expect the frame-level prompts to describe a single corgi in the earlier frames and to have two corgis in the latter frames. This is achieved by leveraging the LLM for complementing the user prompt by accounting for different items we mentioned in Sec.~\ref{sec:motivation}. We show the resultant generated prompts in Fig.~\ref{fig:llm}.

\subsection{Incorporating various contexts with rotational value mapping}

Given frame-level dynamic prompts that account for the story within the video, the remaining challenge for lifting T2I models for T2V is generating frames satisfying the temporal consistency while harmonizing with descriptive frame-level prompts. This requires adjacent frames to have similar time-invariant components, such as object appearances and the background, while still allowing temporal variations, such as movements, to happen. Partially addressing this challenge, T2V-Z~\cite{khachatryan2023text2video} propagates the key and value projection of the first frame of the video across every other frames. However, this constrains the context and content of the overall video to resemble the first frame, without the ability to distinguish time-variant/invariant components. Not only this limits the flexibility of the video, but also dissatisfies the narrative consistency as dynamic motions and transitions can be introduced to the scene through time-variant components.

To overcome these, we introduce value mapping (VM), a method that injects temporal consistency, while enabling the use of diverse contents, such as objects and textures, across the video frames. Different from prior works~\cite{khachatryan2023text2video,qi2023fatezero}, this method adjusts the value of self-attention in relation to the timestep, effectively preventing the objects visual collapsing and ensuring temporal consistency.

Given the formal definition of self-attention layer within a diffusion U-Net~\cite{ho2020denoising,ronneberger2015u}:
\begin{equation}
\textrm{Attention}(Q_{f},K_{f},V_{f})=
\textrm{Softmax}\left(\frac{Q_f (K_{f})^\text{T}}{\sqrt{d}}\right)V_f
\label{eq:self-attention}
\end{equation}
for all frames $f$ in the set $\{1, \dots, F\}$, where $d$ is a channel dimension, and the notation $Q_{f}$, $K_{f}$, and $V_{f}$ is the query, key, and value of the $f$-th frame, respectively. In VM, we modify Eq.~\ref{eq:self-attention} to following:
\begin{align}
\textrm{VM}(Q_{f}, K_{1:F}, V_{1:F}) = \textrm{Softmax}\left(\frac{Q_f (K_{r(t')})^\text{T}}{\sqrt{d}}\right) V_{r(t')}
\label{eq:avm}
\end{align}
for all $f$'s, where $r: \{1, \ldots, T'\} \to \{1, \ldots, F\}$. Here, $t'$ is the number of timesteps from the timestep at which we initiate the value mapping in the diffusion reverse process.

To decide the reference frame $r(t')$ for value mapping, we can simply consider randomly selecting $r(t')$ every timestep for allowing bidirectional flow of value mapping. However, we find this stochastic approach to often result in degenerate results due to the finite number of frames and timesteps. In this regard, we introduce rotational value mapping (RVM), sequentially applying the value of self-attention based on the timestep by rotating over the frames periodically. On a high level, RVM not only enables the complete, bidirectional flow of value mapping but also makes each frame equally contributable. Specifically, we define the reference frame as $r(t') = \textrm{Mod}(\lfloor t'/m\rfloor, F) + 1$, where $m$ represents the period of timesteps. By setting $m$ to a sufficiently large value, RVM becomes equivalent to the cross-frame attention in the original T2V-Z model~\cite{khachatryan2023text2video}. Empirically, setting $m$ to $4$ stabilizes the outcome while maintaining the capability to incorporate various content specified in the frame-level prompts.

\begin{figure*}[!t]
\centering
\includegraphics[width=0.8\linewidth]{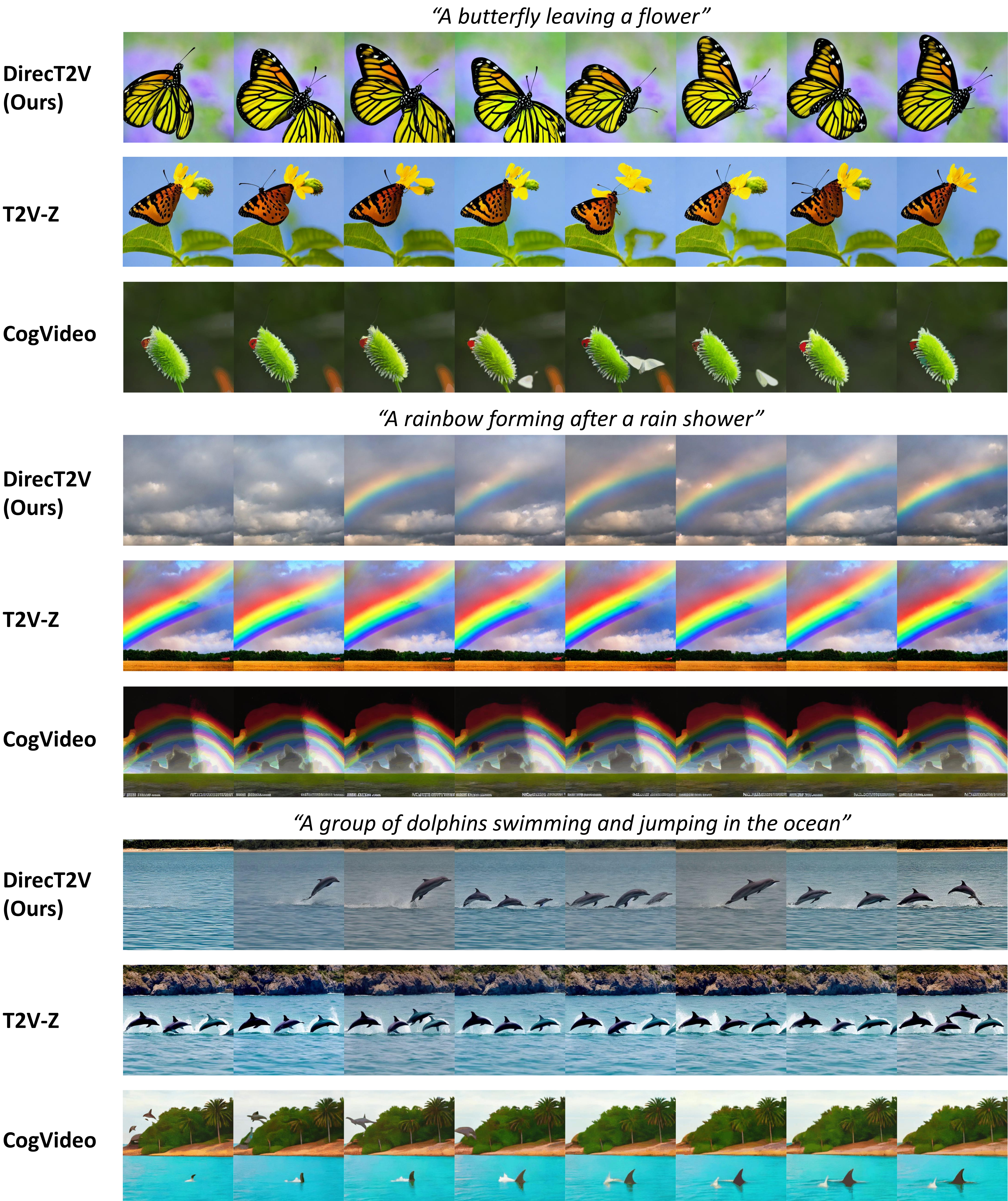}
\vspace{-10pt}
\caption{\textbf{Zero-shot video generation results.} Given an abstract user prompt, we compare DirecT2V with T2V-Z~\cite{khachatryan2023text2video} and CogVideo~\cite{hong2022cogvideo}. 
Note that CogVideo is trained with video-text dataset, instead of zero-shot generation.
}
\label{fig:qual}
\end{figure*}

\begin{figure*}[!t]
\centering
\includegraphics[width=0.8\linewidth]{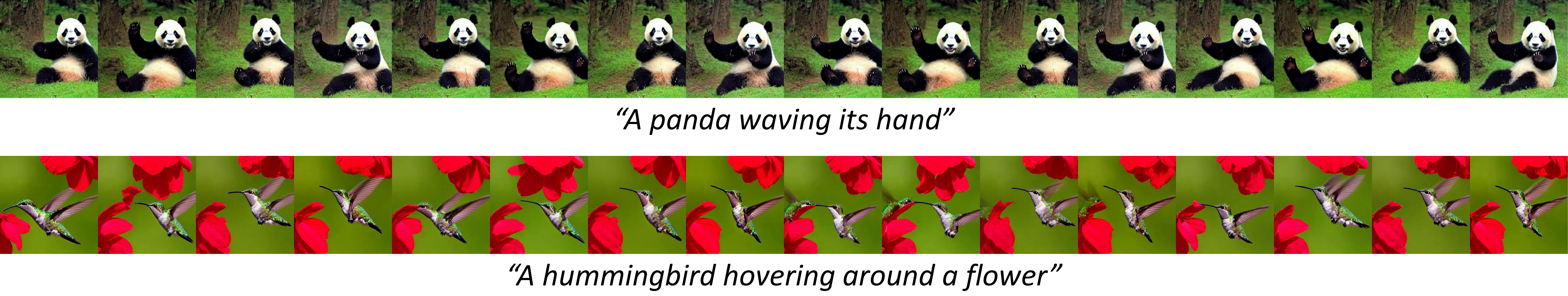}
\vspace{-10pt}
\caption{\textbf{Lifted frame rate.} By iteratively dividing frame-wise prompts, we can generate a video with an arbitrary frame rate.}
\label{fig:lifting}
\vspace{-10pt}
\end{figure*}

\subsection{Reducing unreliable matching via dual softmax filtering}

\begin{figure}[t]
\centering
\includegraphics[width=1.0\linewidth]{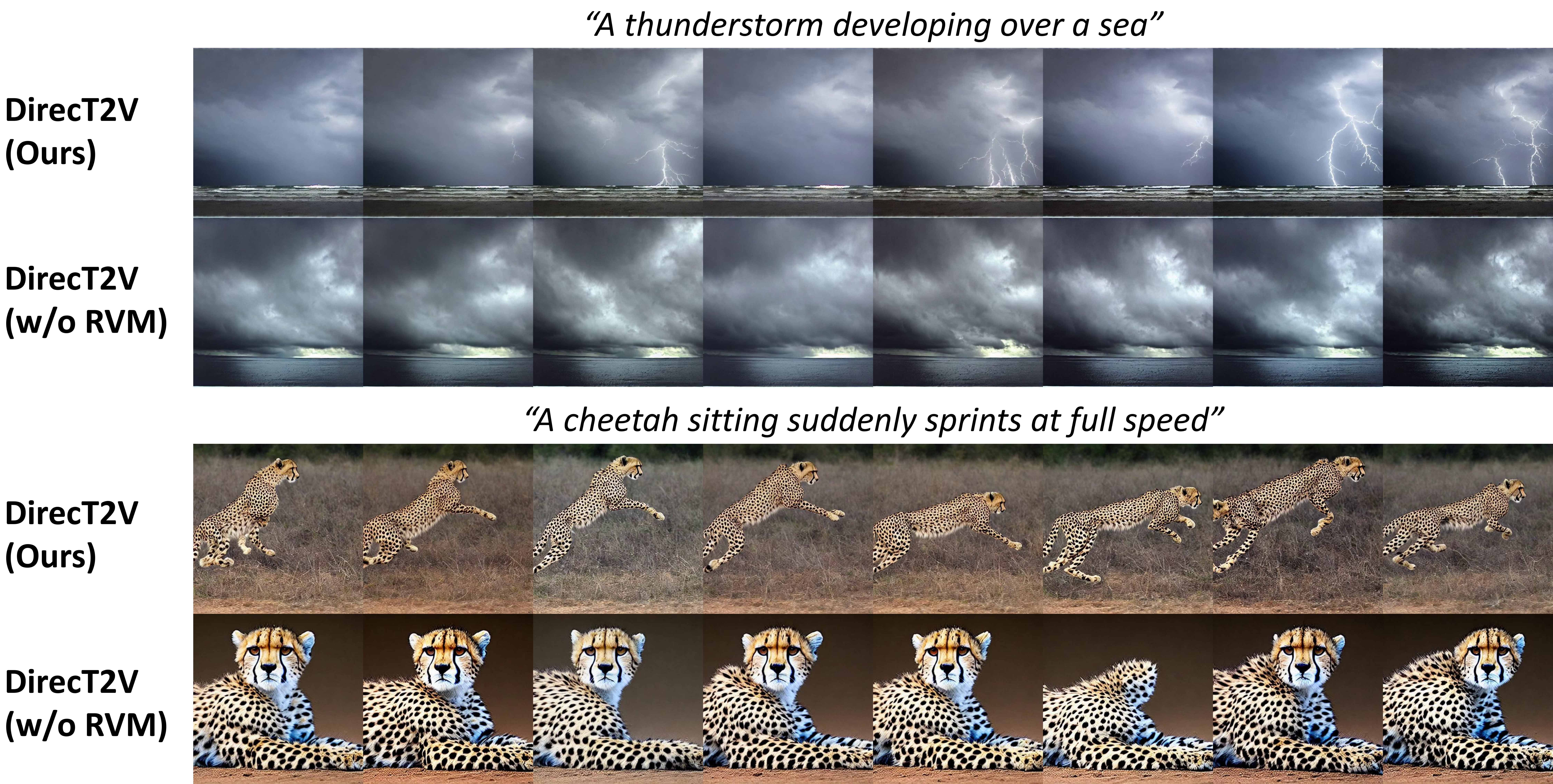}
\vspace{-20pt}
\caption{\textbf{Ablation study on RVM.} With frame-wise directing, RVM is essential for achieving narrative consistency, since its ablated counterpart (without RVM) does not reflect the frame-wise prompt well. Note that the same frame-wise prompts are used for both, while they show different results because RVM mixes various contexts across the frames to better reflect the upsampled prompts.}
\vspace{-10pt}
\label{fig:rotational}
\end{figure}

Although VM can diversify the contextual information within the generated video, on its own, it face difficulties when accounting for rapid movements or drastic changes. In Eq.~\ref{eq:avm}, VM enforces a mapping from $V_{r(t')}$ to $V_f$, where the query-key map $Q_f(K_{r(t')})^{\text{T}}$ can be viewed as a correspondence map~\cite{teed2020raft} that establishes a matching between source frame $r(t')$ and target frame $f$. However, when drastic changes happen between frames, there may exist cases where a reliable matching cannot be established, as an object may not co-occur between the frames. This restricts the target frame from incorporating attributes that are absent in the source frame, thereby preventing desired variations.

To address the issue of unreliable matching, we propose a means for mapping values when a reliable correspondence is established~\cite{mac2012learning,cheng2021improving,truong2021learning, seo2022midms}. This allows us to account for unreliable matching by propagating the original value of the target frame $V_f$ instead of mapping from the reference frame, $V_{r(t')}$. To derive confidence values, we leverage dual softmax~\cite{cheng2021improving} and apply a threshold to these values using a specified quantile. Starting from Eq.~\ref{eq:avm}, the dual softmax~\cite{cheng2021improving}, denoted as $C_\textrm{dual}$, is defined as follows:
\begin{equation}
C_\textrm{dual} = \textrm{Softmax}(Q_t (K_{r(t')})^\text{T}) \odot \textrm{Softmax}(K_{r(t')} (Q_t)^\text{T}),
\label{eq:dual-softmax}
\end{equation}
where $\odot$ represents the Hadamard product. This is followed by masking to map only the reliable values:
\begin{align}
\begin{split}
&\textrm{VM}'(Q_{f}, K_{1:F}, V_{1:F}) \\ &= (1-M_f) \odot \textrm{Attention}(Q_f, K_f, V_f)\\&+ M_f\odot\textrm{VM}(Q_{f}, K_{1:F}, V_{1:F})
\label{eq:reliable}
\end{split}
\end{align}
for all $f$'s, where $M_f = \mathbbm{1}(\mathcal{A}(C_\textrm{dual}) > \phi)$ for an averaging and broadcasting operation denoted as $\mathcal{A}(\cdot)$, and $\phi$ is a pre-defined quantile of $\mathcal{A}(C_\textrm{dual})$.
This method allows only confident inter-frame matching, reflecting desired variances while preventing distortion throughout the video sequence.

\section{Experiments}

\subsection{Implementation details}

In this work, we employ GPT-4~\cite{openai2023gpt4} as our instruction-tuned LLM and T2V-Z~\cite{khachatryan2023text2video}, utilizing a single NVIDIA 3090 RTX GPU for efficient video sampling. For the generation process, we employ the PNDM scheduler~\cite{liu2022pseudo}, which is a member of the deterministic diffusion samplers family~\cite{karras2022elucidating,song2021denoising,liu2022pseudo}. We employ the same scheduling mechanism as T2V-Z~\cite{khachatryan2023text2video} and configure the scheduler parameters of the diffusion models with $T=100$ and $T'=96$. Furthermore, we adopt the classifier-free guidance~\cite{ho2021classifier}, using a scale of $12.0$. Both T2V-Z and our method employ motion dynamics; however, for a fair comparison, we refrain from using it in the main paper by setting its intensity to zero.% We visualize results with motion dynamics in the supplemental material. The code will be made publicly available.

\subsection{Zero-shot video generation results}

\paragraph{Qualitative results.}
In Fig.~\ref{fig:qual}, we showcase the zero-shot video generation capabilities of DirecT2V. Our framework generates per-frame prompts based on a user's description of a general scene, and with our attention mechanism, the prompts are animated into a video containing dynamic actions and time-varying content.

\begin{table}[t]
\scriptsize
\centering
\begin{tabular}{c|cccc}
\toprule
Frame \# & T2V-Z & TAV & RVM & RVM+DSF \\
\midrule
1 & 0.3208 & 0.3236 & 0.3143 & 0.3180 \\
2 & 0.2950 & 0.2947 & 0.3026 & 0.3077 \\
3 & 0.2894 & 0.2909 & 0.3056 & 0.3077 \\
4 & 0.2865 & 0.2931 & 0.3123 & 0.3095 \\
5 & 0.2935 & 0.3006 & 0.3137 & 0.3131 \\
6 & 0.2889 & 0.3013 & 0.3052 & 0.3122 \\
7 & 0.2858 & 0.2980 & 0.3103 & 0.3142 \\
8 & 0.2888 & 0.2988 & 0.3052 & 0.3077 \\
\midrule
Avg. & 0.2930 & 0.3001 & 0.3087 & \textbf{0.3113} \\
Avg. Dist. & 0.0272 & 0.0235 & \textbf{0.0057} & 0.0067 \\
\bottomrule
\end{tabular}
\vspace{-5pt}
\caption{\textbf{Comparison of various attention mechanisms.}}
\vspace{-10pt}
\label{tab:quan}
\end{table}

From the results, given a prompt \textit{``A corgi is running and another corgi joins later,''} DirecT2V successfully portrays the second corgi appearing in the intermediate frames. In contrast, the second corgi is either always present or entirely absent for T2V-Z. As shown for the other prompt, \textit{``A rainbow forming after a rain shower,''} demonstrates our method's ability to synthesize videos with narrative consistency, whereas T2V-Z with just the user prompt shows the rainbow exists for every frame.

Notably, the generated video from CogVideo~\cite{hong2022cogvideo} also lacks some components of the given user prompts, even though the model is fine-tuned using text-video pairs. For example, the addition of another corgi or a change in weather is not incorporated into the generated videos. These results corroborates the effectiveness of our frame-wise prompting approach.

\vspace{-5pt}
\paragraph{Quantitative results.}

Although our zero-shot lifting from image diffusion models does not have a ground-truth video reference dataset, several works use the CLIP score as a metric to evaluate faithfulness. Following this, we display in Table~\ref{tab:quan} the CLIP similarity score comparison for each frame. For the comparison, we used the proposed attention mechanism in Text2Video-Zero (T2V-Z)~\cite{khachatryan2023text2video}, Tune-A-Video (TAV)~\cite{wu2022tune}, our rotational value mapping, and rotational value mapping with dual softmax filtering. Note that for TAV, we only leveraged the attention mechanism for zero-shot video generation. In terms of average similarity, our attention mechanism clearly outperforms the others. \textit{Avg. Dist.} in this table refers to the average L1 distance between frame 1 and the other frames, and we can observe that both T2V-Z and TAV are only faithful to the first prompt.

\vspace{-5pt}

\begin{table}[t]
\small
\centering
\begin{tabular}{lccc}
\toprule
Method & Faithfulness & Realism & Narrative \\
\midrule
T2V-Z & 9.52\% & 6.55\% & 7.14\% \\
Ours & \textbf{90.48}\% & \textbf{93.45}\% & \textbf{92.86}\% \\
\bottomrule
\end{tabular}
\vspace{-5pt}
\caption{\textbf{User study.}}
\vspace{-15pt}
\label{tab:user}
\end{table}

\vspace{-5pt}
\paragraph{User study}

We also conducted a user study on the faithfulness to the user prompt, realism, and narrative of the generated videos, comparing them with the state-of-the-art zero-shot text-to-video baseline~\cite{khachatryan2023text2video}. The results are shown in Table~\ref{tab:user}, which demonstrate that ours is preferable to human evaluators.

\subsection{Controlling video attributes with LLMs}

In this section, we show that by providing specific instructions to LLMs, we can control the attributes of the video, such as the number of frames and frames per second (FPS). For lifting FPS, we provide instructions to the LLM to divide the prompts, and use it to generate videos similar to the method described above. This approach effectively handles situations where the number of frames exceeds the batch size. For controlling attributes other than FPS, we provide further results in the appendix.

Originally, the frame-wise director extracts $F$ frames at a specified frame rate $R$. We then provide the following prompt: \textit{``Now, at a frame rate of} \{$2 \times R$\} \textit{fps, divide each frame in the previous result into two separate image descriptions. This should eventually result in} \{$2 \times F$\} \textit{frames.''} We repeat this process with the new $F:=2F$.

For the caching process, if the number of frames exceeds the batch size, we divide it into sections with a size of floored half of the batch size. Initially, we perform rotational value mapping for the frames, the number of which is the floored half of the batch size. Subsequently, we utilize the cached attention from the first process to perform rotational value mapping for the full batch size. The results are displayed in Fig.~\ref{fig:lifting}.

\subsection{Ablation study}

\paragraph{Rotational value mapping.}
In Fig.~\ref{fig:rotational}, we display the ablation study for our Rotational Value Mapping (RVM) approach. This experiment showcases the generated video outcomes alongside the results obtained without employing rotational selection, \textit{i.e.}, utilizing only the keys and values of the initial frame, as seen in \cite{khachatryan2023text2video}.

When given a prompt like \textit{``A thunderstorm developing over a sea,''} DirecT2V effectively depicts the thunderstorm's subsequent progression, as described by the GPT-4-generated prompt for that specific frame. On the other hand, the ablated version, with a fixed value in the second row, fails to develop a thunderstorm using the given fixed value. For another prompt, \textit{``A cheetah sitting suddenly sprints at full speed,''} our technique illustrates its capacity to create videos that maintain narrative coherence. In contrast, DirecT2V without RVM leads to a persistent first-frame prompt. These findings emphasize RVM's ability to generate videos featuring dynamic actions and preserved time-varying content.

\vspace{-5pt}
\paragraph{Dual-softmax filtering.}

\begin{figure}[t]
\centering
\includegraphics[width=1.0\linewidth]{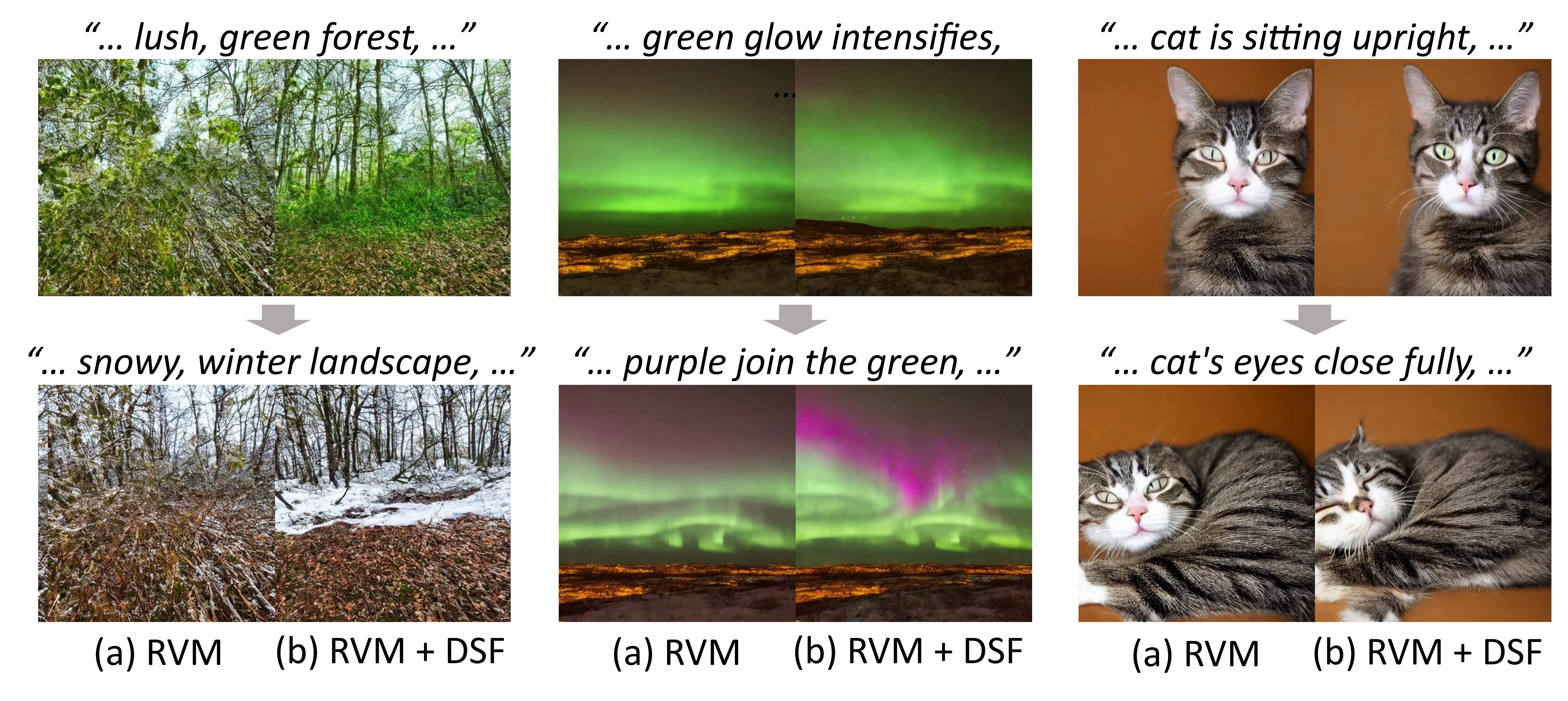}
\vspace{-20pt}
\caption{\textbf{Ablation study on DSF.} For each prompts, results on the left are only with RVM, while the right are with RVM and Dual-Softmax Filtering (DSF).}
\vspace{-15pt}
\label{fig:dual-softmax}
\end{figure}

In the experiment depicted in Fig.~\ref{fig:dual-softmax}, we present the outcomes obtained when applying dual-softmax filtering, which successfully addresses the negative impact of inaccurate matches. For this study, we select two separate frames from the generated video and exhibit them along with the prompt proportion. It is worth noting that in the third column, the absence of dual-softmax filtering results in inevitable inaccurate matching due to value mapping, leading to the cat's eyes from a different frame being mapped onto the closed eyes.

\section{Conclusion}

In this paper, we present a novel approach for zero-shot video creation from textual prompts, tackling the intricate challenges of maintaining temporal consistency and visual quality in the generated videos. By employing GPT-4, a state-of-the-art instruction-tuned language model, we demonstrate its capability to generate detailed and temporally consistent image descriptions, which can be effectively integrated into text-to-image models. We introduce two key innovations for frame interactions, rotational value mapping and dual softmax filtering, which significantly enhance the flexibility and overall quality of the generated videos. Our method surpasses others in both the composition of the storyline and quality.

\section*{Impact Statements}
This paper presents work whose goal is to advance the field of Machine Learning. There are many potential societal consequences of our work, none which we feel must be specifically highlighted here.

% In the unusual situation where you want a paper to appear in the
% references without citing it in the main text, use \nocite
% \nocite{langley00}

\bibliography{main}
\bibliographystyle{icml2024}

%%%%%%%%%%%%%%%%%%%%%%%%%%%%%%%%%%%%%%%%%%%%%%%%%%%%%%%%%%%%%%%%%%%%%%%%%%%%%%%
%%%%%%%%%%%%%%%%%%%%%%%%%%%%%%%%%%%%%%%%%%%%%%%%%%%%%%%%%%%%%%%%%%%%%%%%%%%%%%%
% APPENDIX
%%%%%%%%%%%%%%%%%%%%%%%%%%%%%%%%%%%%%%%%%%%%%%%%%%%%%%%%%%%%%%%%%%%%%%%%%%%%%%%
%%%%%%%%%%%%%%%%%%%%%%%%%%%%%%%%%%%%%%%%%%%%%%%%%%%%%%%%%%%%%%%%%%%%%%%%%%%%%%%
\newpage
\appendix
\onecolumn
\section{Zero-shot video generation results with motion dynamics}

\begin{figure}[t]
\centering
\includegraphics[width=0.85\linewidth]{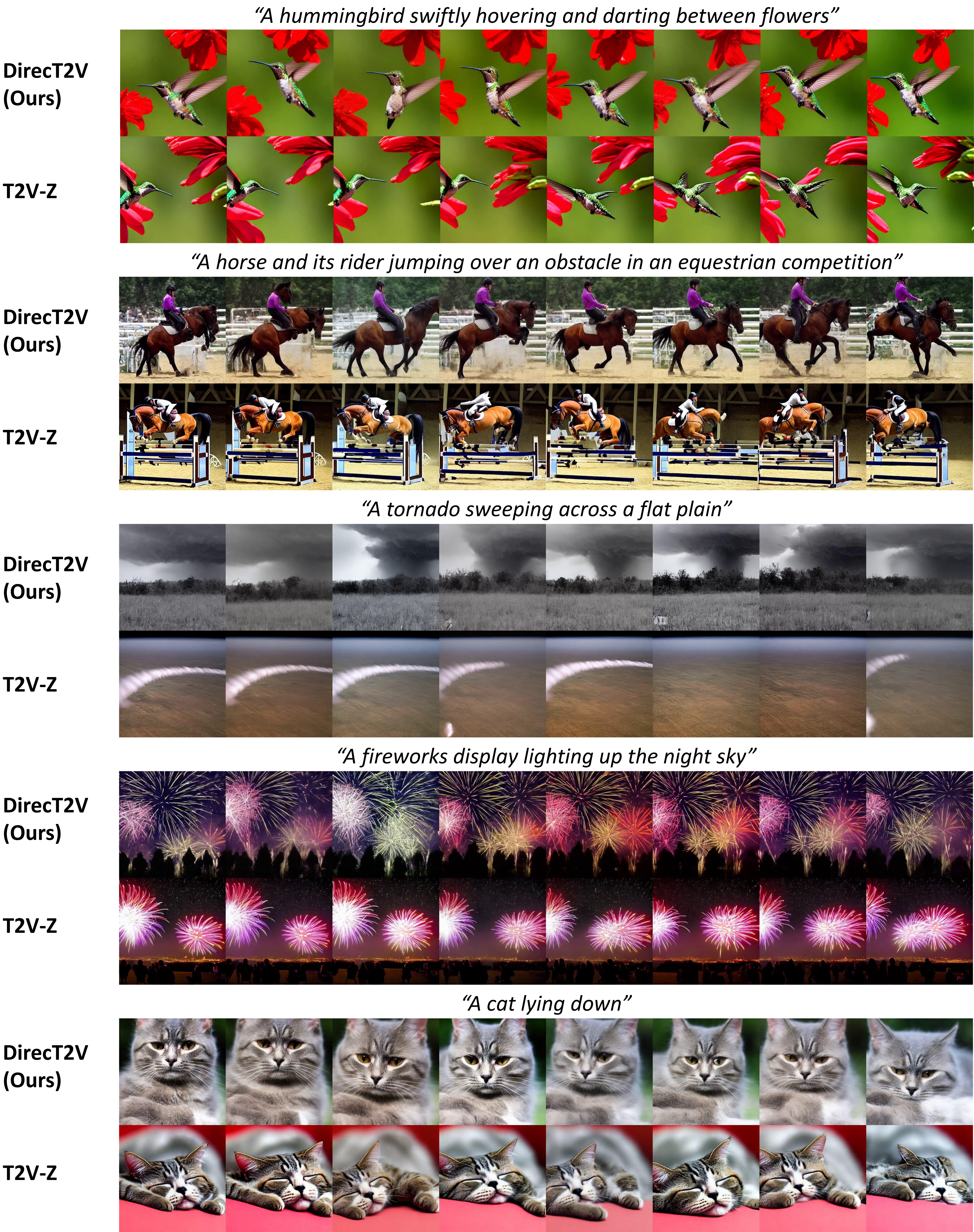}
\caption{\textbf{Zero-shot video generation results with motion dynamics~\cite{khachatryan2023text2video}.} We compare our method with T2V-Z~\cite{khachatryan2023text2video}.}
\label{fig:abl-qual}
\end{figure}

\begin{figure}[t]
\centering
\includegraphics[width=0.85\linewidth]{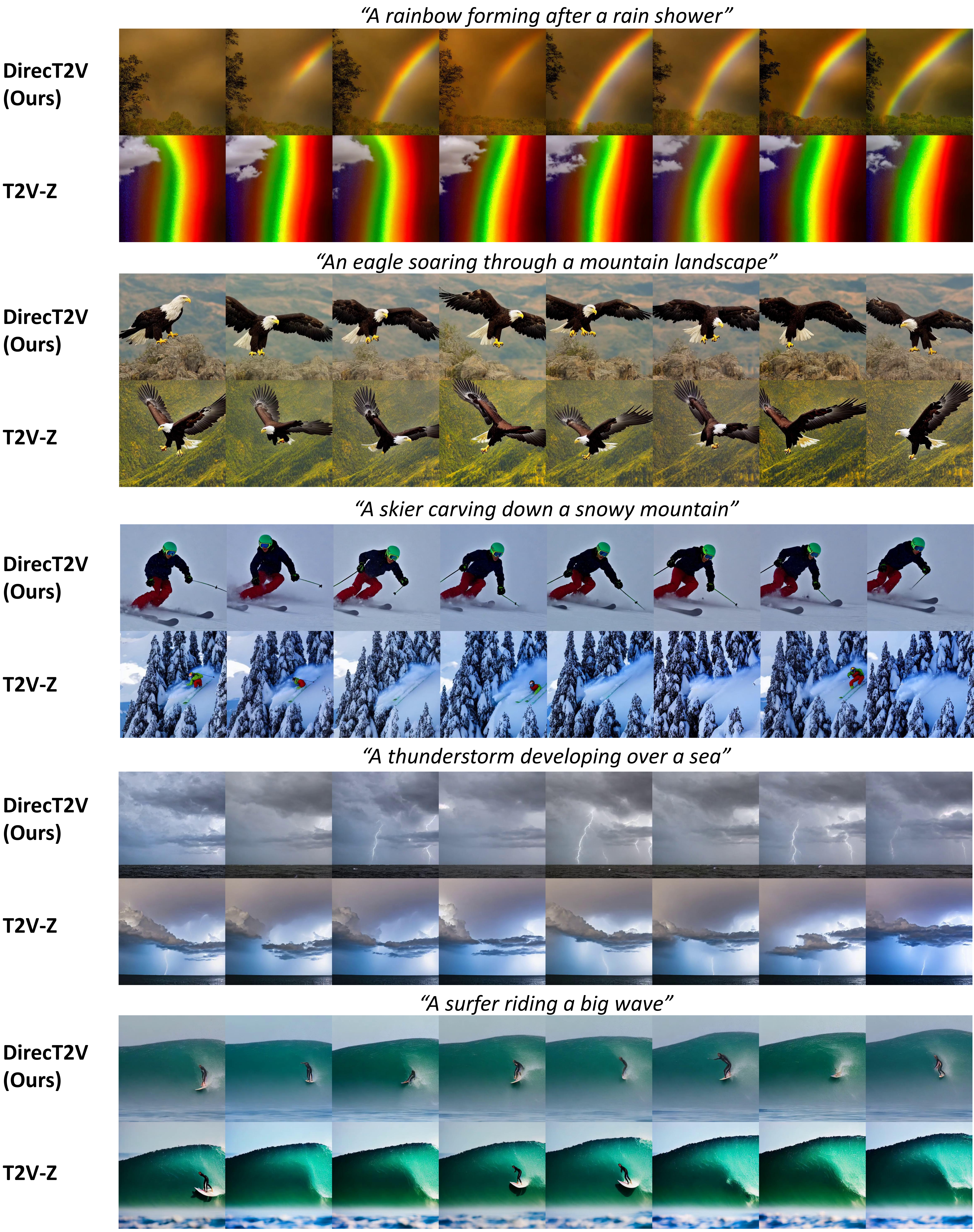}
\caption{\textbf{Zero-shot video generation results with motion dynamics~\cite{khachatryan2023text2video}.} We compare our method with T2V-Z~\cite{khachatryan2023text2video}.}
\label{fig:abl-qual2}
\end{figure}

Our framework can naturally be extended to provide motion dynamics~\cite{khachatryan2023text2video}, a feature that enables the capturing of explicit camera movement, \textit{i.e.}, translation. Given accurately predicted motion dynamics, our approach would not only encapsulate the context of the narrative but also dynamically move with it. The results are provided in Fig.~\ref{fig:abl-qual} and Fig.~\ref{fig:abl-qual2}, and the video is available on our homepage.

\section{Experimental setting}

\begin{figure}[t]
\centering
\includegraphics[width=0.8\linewidth]{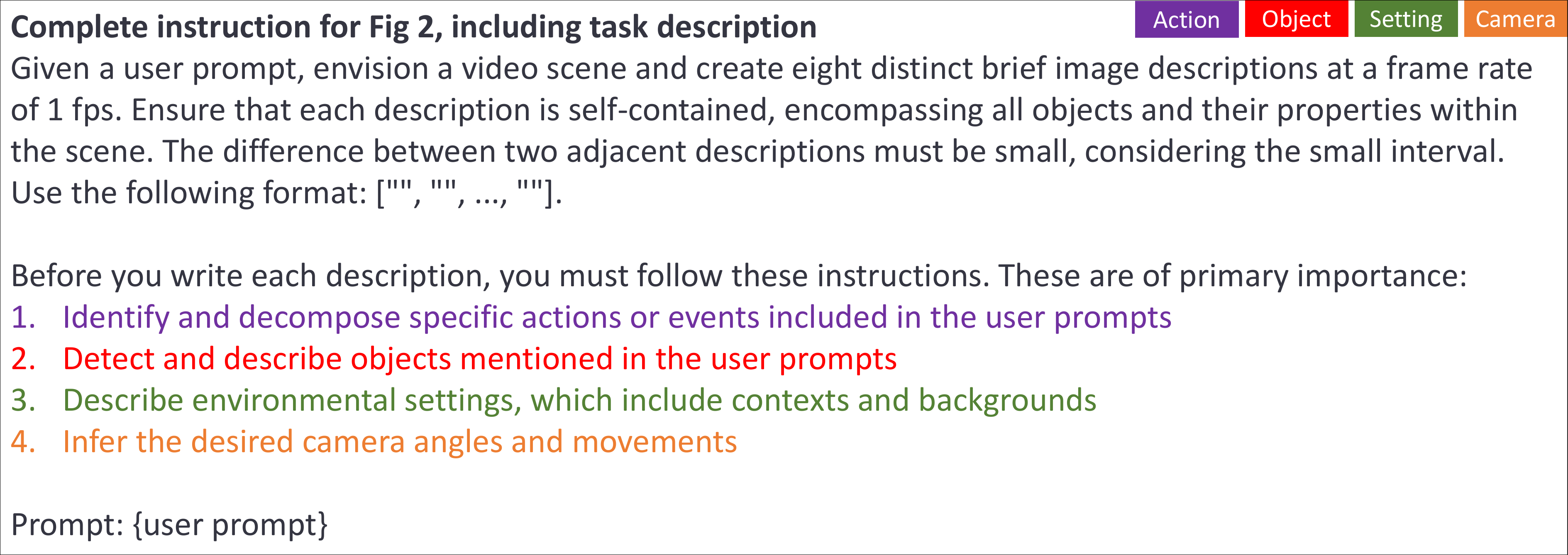}
\caption{\textbf{The complete instruction for Fig.~\ref{fig:llm}, including the task description.}}
\label{fig:full-prompt}
\end{figure}

\begin{figure}[t]
\centering
\includegraphics[width=1.0\linewidth]{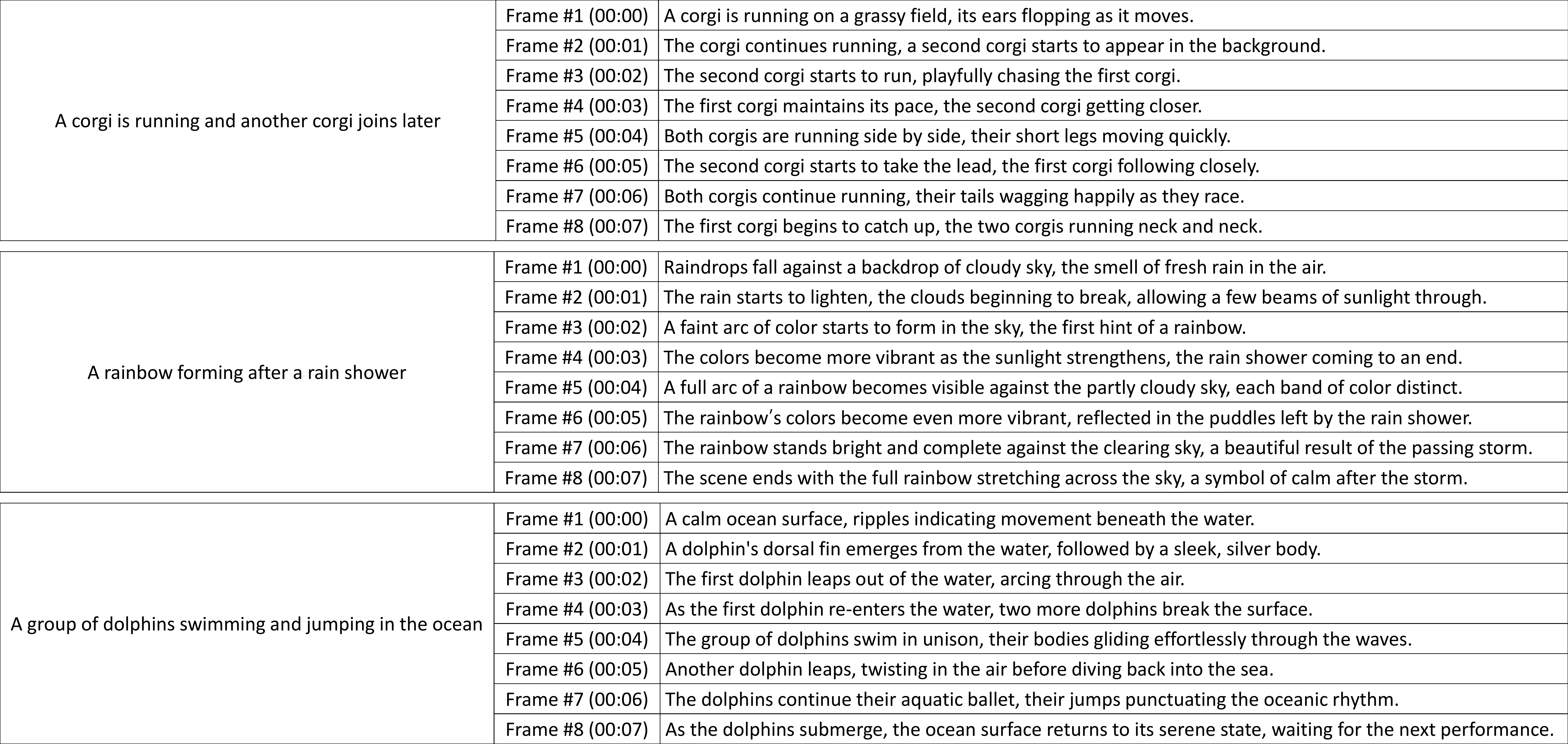}
\caption{\textbf{Examples of frame-level prompts directed by LLMs.}}
\label{fig:framewise}
\end{figure}

We present examples of the full prompts used to produce the results shown in Fig.~\ref{fig:llm} of the main paper. These examples are depicted in Fig.~\ref{fig:full-prompt}. Additionally, in Fig.~\ref{fig:framewise}, we provide examples of the frame-wise prompts utilized in creating videos.

\section{More ablation studies and analyses}

\begin{figure}[t]
\centering
\includegraphics[width=1.0\linewidth]{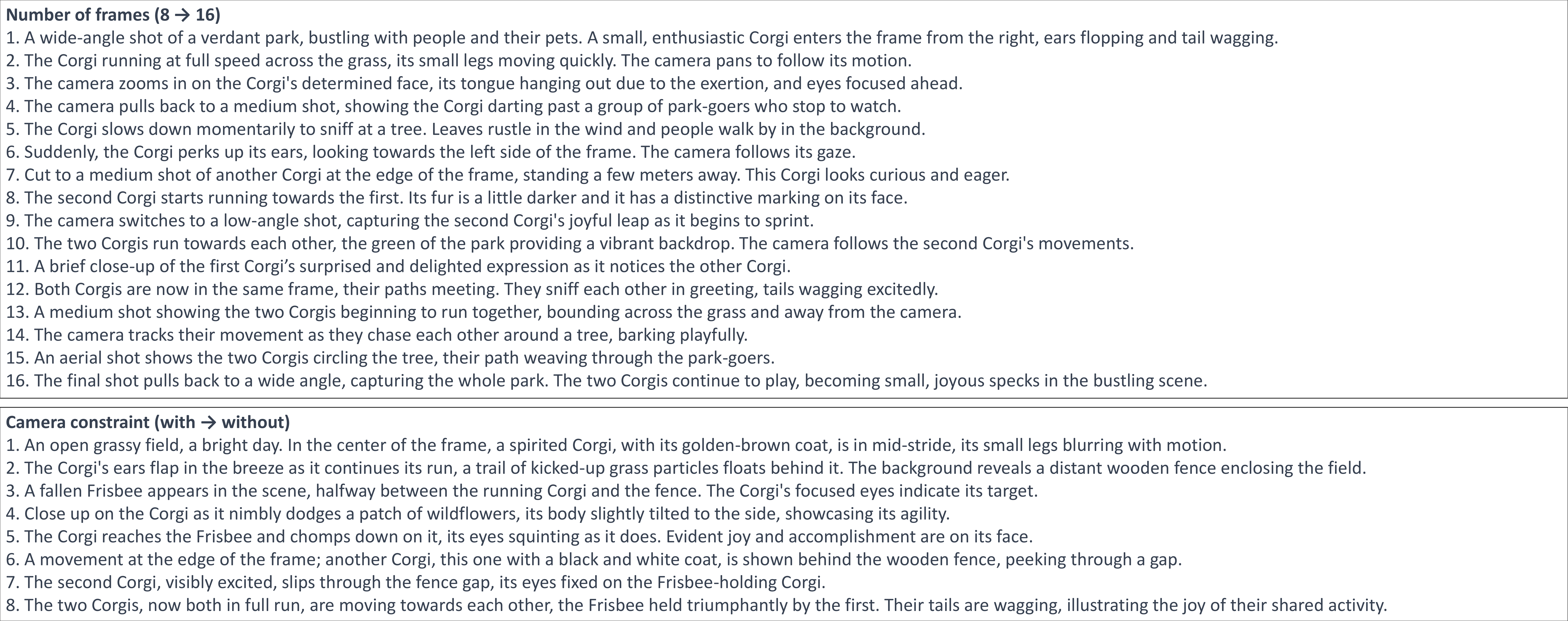}
\caption{\textbf{Video attribute control using LLMs.}}
\label{fig:attribute}
\end{figure}

\paragraph{Attribute controls.} In the main paper, we discuss the effects of regulation of frame rate control. In this section, we also provide our results for attribute control. These results are presented in Fig.~\ref{fig:attribute}. In this experiment, we demonstrate the capability to manipulate the number of frames effectively, as well as to remove or add the camera setting of the video scene from a simple prompt.

\paragraph{Rotational and random value mapping.}

\begin{wrapfigure}{r}{7.0cm}
\vspace{-10pt}
\centering
\includegraphics[width=1.0\linewidth]{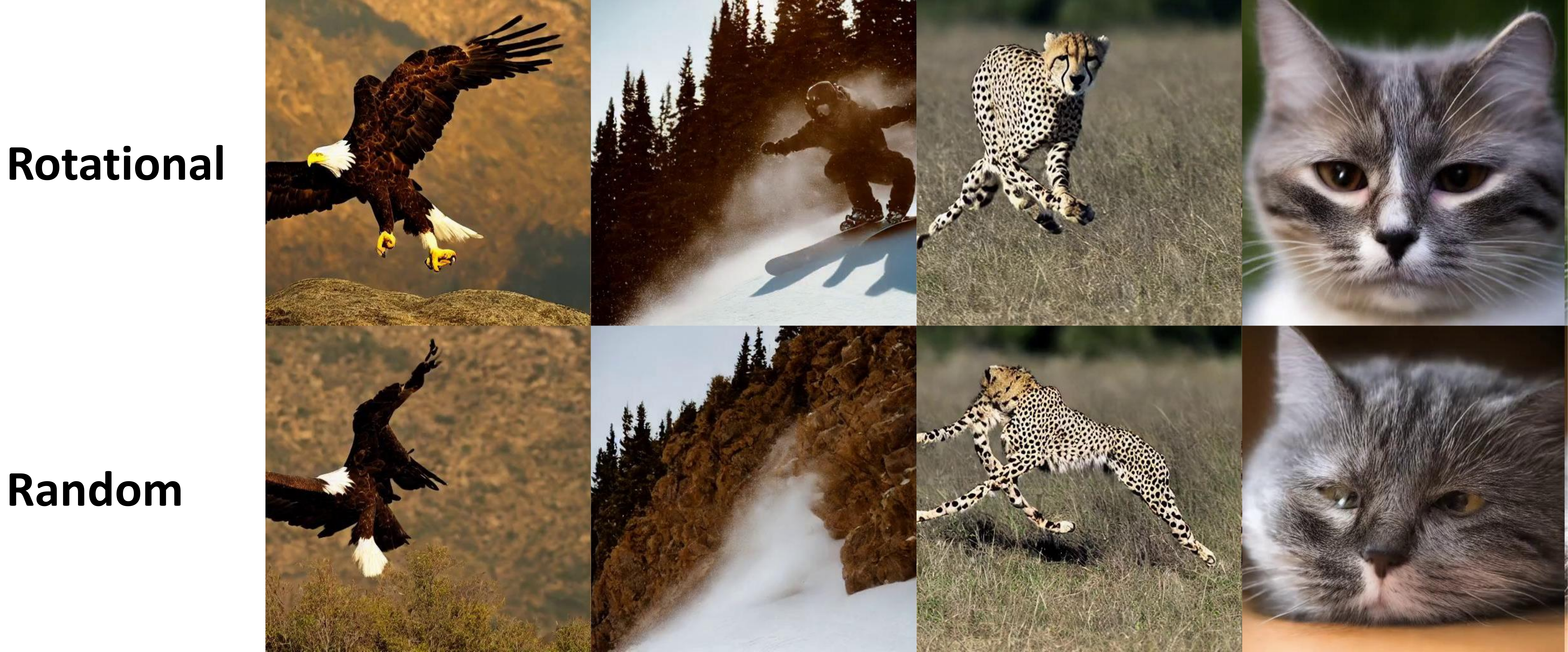}
\caption{\textbf{Examples of results from the different value mappings.}}
\vspace{-30pt}
\label{fig:avm-comp}
\end{wrapfigure}

We also test the random method of value mapping, which can replace our RVM. This means the reference frames are randomly selected contrary to RVM. The results, however, show that the method adopting random mapping obtains results with occasionally distorted objects, possibly due to the instability caused by the small number of timesteps.

\paragraph{User study.}
In each case, two videos were provided, showing our results compared to the baseline. The identity of the videos (such as which video is ours) was completely concealed, and the placement of the videos was also randomized. This questionnaire was distributed over three days to local communities and universities, and stakeholders of the study were strictly excluded. We aggregated 168 answers from 21 participants, each of whom was asked to answer three questions about eight assigned cases. The questions were as follows:
\begin{itemize}
    \item The videos in the first and second rows are the results generated from different models.
    \begin{itemize}
    \item Which video is better aligned with the text displayed at the top of the videos?
    \item Which video appears more realistic and natural?
    \item Which video better demonstrates the narrative given in the text?
    \end{itemize}
\end{itemize}

\paragraph{Comparison of GPT-4, GPT-3.5, and Bard.}
We compare the frame-level directing abilities of GPT-4~\cite{openai2023gpt4}, GPT-3.5~\cite{openai2022chatgpt}, and Bard~\cite{google2023palm2}. In Fig.~\ref{fig:exp-prompt}, we present the conversations with those LLMs. We find that they generally follow the instruction prompt well.

\section{Discussion}
\paragraph{Concurrent work.}
In a short time since this paper, there have been a large number of works on text-to-video using image or video diffusion models~\cite{huang2023free,blattmann2023align,esser2023structure,guo2023animatediff,an2023latent,luo2023videofusion,chen2023control,li2023videogen}. Among these, there is a concurrent endeavor that also leverages LLMs for zero-shot text-to-video generation, Free-Bloom~\cite{huang2023free}, which serves as supporting evidence for our motivation to use LLMs in this task. Free-Bloom adopts the previously known attention mechanism, which we have experimented with in our ablation study, with the key difference being its cessation of references to other frames at a timestep. Furthermore, it employs a different noise sampling technique. Overall, their work is mostly orthogonal to ours in terms of the attention mechanism and video generation.

\begin{figure}[t]
\centering
\includegraphics[width=1.0\linewidth]{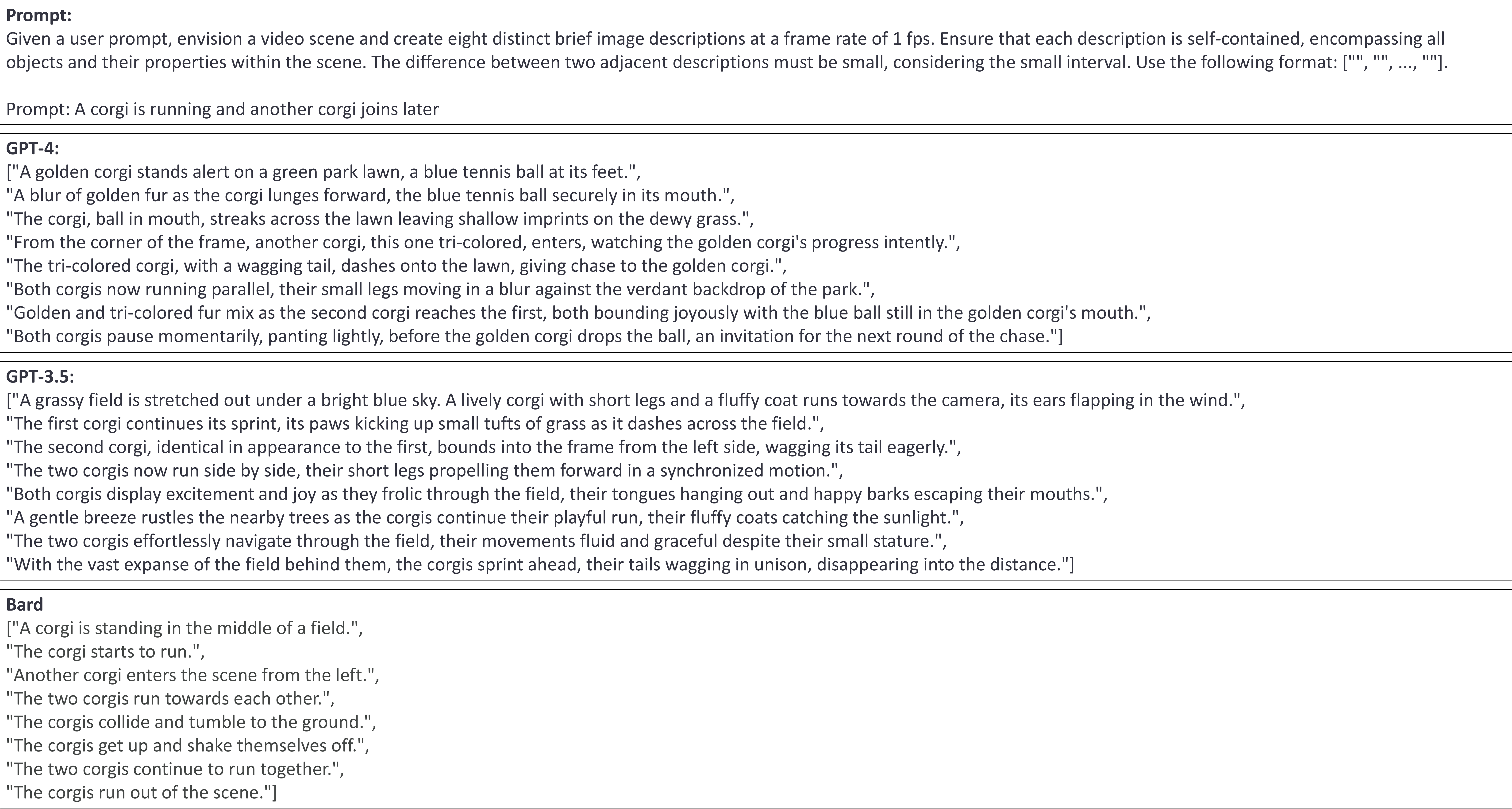}
\caption{\textbf{Comparison between frame-level prompting of GPT-4~\cite{openai2023gpt4}, GPT-3.5~\cite{openai2022chatgpt}, and Bard~\cite{google2023palm2}.}}
\label{fig:exp-prompt}
\end{figure}

\paragraph{Limitations.}
The performance of the proposed method, DirecT2V, may vary depending on the instruction-tuned large language models (LLMs)~\cite{openai2023gpt4,openai2022chatgpt,google2023palm2} (see Fig.~\ref{fig:exp-prompt}). As a result, any biases or limitations within these models may adversely affect the quality of the resulting videos. This is because LLMs can produce ambiguous or distracting descriptions, leading to less accurate or coherent video frames. Further research might explore the incorporation of additional constraints to create more vision-friendly frame-by-frame prompts. Moreover, DirecT2V's dependence on pre-trained text-to-image diffusion models introduces another layer of dependency. These models have encountered challenges in accurate counting and positioning~\cite{saharia2022photorealistic,li2023gligen}. A potential solution to this problem could involve the use of the encoder from an even larger language model~\cite{saharia2022photorealistic}.

In essence, the development and enhancement of both instruction-tuned LLMs and T2I diffusion models equipped with attention mechanisms present promising landmarks for the future improvement of DirecT2V.

\paragraph{Broader impact.}
To the best of our knowledge, our research presents DirecT2V as the first framework that explicitly leverages the temporal and narrative knowledge embedded within large language models for high-level visual creation, specifically video creation~\cite{khachatryan2023text2video,singer2022make,hong2022cogvideo,singer2022make,ho2022imagen,villegas2022phenaki,wu2022nuwa,zhou2022magicvideo}. This philosophy can be extended to other high-level visual tasks, such as zero-shot text-to-3D~\cite{hong2023debiasing,poole2022dreamfusion,wang2022score,metzer2022latent,lin2022magic3d,chen2023fantasia3d,seo2023let,xu2022dream3d,seo2023ditto}. Nevertheless, the ability of DirecT2V to generate realistic videos from textual prompts raises concerns about its potential to contribute to the spread of misinformation and deepfake content. The increased difficulty in distinguishing between authentic and fabricated videos may exacerbate existing concerns about the dissemination of false information.

%%%%%%%%%%%%%%%%%%%%%%%%%%%%%%%%%%%%%%%%%%%%%%%%%%%%%%%%%%%%%%%%%%%%%%%%%%%%%%%
%%%%%%%%%%%%%%%%%%%%%%%%%%%%%%%%%%%%%%%%%%%%%%%%%%%%%%%%%%%%%%%%%%%%%%%%%%%%%%%

\end{document}